\documentclass{article}
\usepackage{spconf,amsmath,graphicx,hyperref}
\usepackage{stfloats}
\usepackage{booktabs} 
\usepackage{multirow}
\usepackage{cite}
\usepackage{amsmath,amssymb,amsfonts}
\usepackage{algorithmic}
\usepackage{graphicx}
\usepackage{textcomp}
\usepackage{xcolor}
\usepackage{booktabs}
\usepackage{multirow}
\usepackage{romannum}
\usepackage{verbatim}
\usepackage{pifont}

\title{MIDG: Mixture of Invariant Experts with knowledge injection for Domain Generalization in Multimodal Sentiment Analysis}
\name{Yangle Li$^{1}$ \qquad Danli Luo$^{1}$ \qquad Haifeng Hu$^{1}$$^{*}$ \thanks{* Corresponding Author}}
  
 \address{$^{1}$ School of Electronics and Information Technology, Sun Yat-Sen University \\}

\begin{document}
\ninept
\maketitle
\begin{abstract}

Existing methods in domain generalization for Multimodal Sentiment Analysis (MSA) often overlook inter-modal synergies during invariant features extraction, which prevents the accurate capture of the rich semantic information within multimodal data. Additionally, while knowledge injection techniques have been explored in MSA, they often suffer from fragmented cross-modal knowledge, overlooking specific representations that exist beyond the confines of unimodal. To address these limitations, we propose a novel MSA framework designed for domain generalization. Firstly, the framework incorporates a Mixture of Invariant Experts model to extract domain-invariant features, thereby enhancing the model's capacity to learn synergistic relationships between modalities. Secondly, we design a Cross-Modal Adapter to augment the semantic richness of multimodal representations through cross-modal knowledge injection. Extensive domain experiments conducted on three datasets demonstrate that the proposed MIDG achieves superior performance.

\end{abstract}
\begin{keywords}
Multimodal Sentiment Analysis, Domain Generalization , Knowledge Injection
\end{keywords}
\section{Introduction}
\label{sec:intro}

Sentiment analysis is a fundamental task in the field of Natural Language Processing (NLP). With the rapid development of multimedia technology, artificial intelligence is gradually entering the multimodal era. As a result, researchers in sentiment analysis have shifted their focus from text-based approaches to Multimodal Sentiment Analysis (MSA), which integrates information from multiple dimensions such as text, audio, and visual signals to analyze human sentiment expressions.

Most existing MSA methods rely on an unrealistic assumption that training and testing data are drawn from an independent identical distribution. However, this assumption rarely holds in real-world scenarios. To mitigate this limitation, previous studies employed sparse masking techniques and extracted invariant features using a text-first-then-video strategy\cite{learning_in_order}, achieving improved generalization performance. Nevertheless, such methods suffer from issue \textbf{(i)}: the sequential processing of modalities fails to adequately model and leverage the dynamic synergy and complementary relationships among textual, acoustic, and visual signals, thus limiting accurate sentimental analysis.

Furthermore, many existing MSA approaches predominantly rely on general knowledge from pre-trained models such as BERT\cite{bert} to encode each modality, which is insufficient to capture cross-modal sentiment specific cues. A potential solution is knowledge injection\cite{survey2}. Prior work\cite{yu2023conki} introduced a multimodal Adapter that leverages pan-knowledge within a modality to generate knowledge-specific representations and inject them into pan-knowledge to facilitate prediction. However, this approach suffers from issue \textbf{(ii)}: current Adapters focus on intra-modal learning and injection, leading to fragmented cross-modal knowledge. They often overlook extra-modal knowledge representations, making it difficult to capture comprehensive semantic information.

To address these two issues, we propose a MoIE with Knowledge Injection for Domain Generalization Framework in MSA (MIDG), which leverages a Mixture of Invariant Experts (MoIE) to learn in-domain invariant features, effectively capturing multimodal collaborative information. Additionally, we introduce a Cross-Modal Adapter to efficiently learn cross-modal specific knowledge representations, thereby enhancing out-of-domain generalization and improving the predictive accuracy of the MSA model. Specifically, to tackle issue \textbf{(i)}, we propose a Mixture of Invariant Experts (MoIE) module to extract invariant features. A router network dynamically assigns tasks to experts based on input features. To address issue \textbf{(ii)}, we simulate out-of-domain information using a subset of raw data to train the model's generalization capability. For this purpose, we design a Cross-Modal Adapter for knowledge injection to produce knowledge-specific representations. We further employ an attention mechanism to dynamically integrate specific knowledge from other modalities.The main contributions of our work are summarized as follows:

\textbf{·}We propose a novel MoIE based knowledge injection domain generalization framework for MSA, which specifically addresses the challenge of extracting invariant features under out-of-distribution scenarios.

\textbf{·}The Mixture of Invariant Experts module is employed to learn multimodal domain-invariant features. Taking the multimodal fused features as input, a routing network dynamically assigns learning tasks to experts, thereby enhancing the learning of cross-modal interactive knowledge.

\textbf{·}To fully leverage inter-modal interactions, a cross-modal knowledge injection module (CM-Adapter) is introduced. It dynamically adjusts the weights of information from different modalities through a multi-head attention mechanism, holistically injecting information from other modalities into the unimodal representations.

\textbf{·}We conducted extensive experiments on the model to demonstrate its performance. Evaluations show that our approach achieves generalization competitive performance compared to state-of-the-art methods.

\section{related work}
\label{sec:format}

\subsection{Multimodal Sentiment Analysis}
\label{ssec:subhead}

Multimodal sentiment analysis can extract richer and more complementary sentimental cues from multimodal information, thereby achieving more robust and accurate sentiment recognition. Research on MSA methods primarily focuses on two categories: multimodal representation learning\cite{confede} and multimodal fusion\cite{fuse}. Methods of representation learning decouple multimodal features into modality-invariant and modality-specific representations to address multimodal heterogeneity. Research on multimodal fusion methods aims to comprehensively integrate multimodal data for MSA tasks, encompassing both feature-level fusion and decision-level fusion.

\begin{figure*}[t]
\centering
\centerline{\includegraphics[scale=0.5]{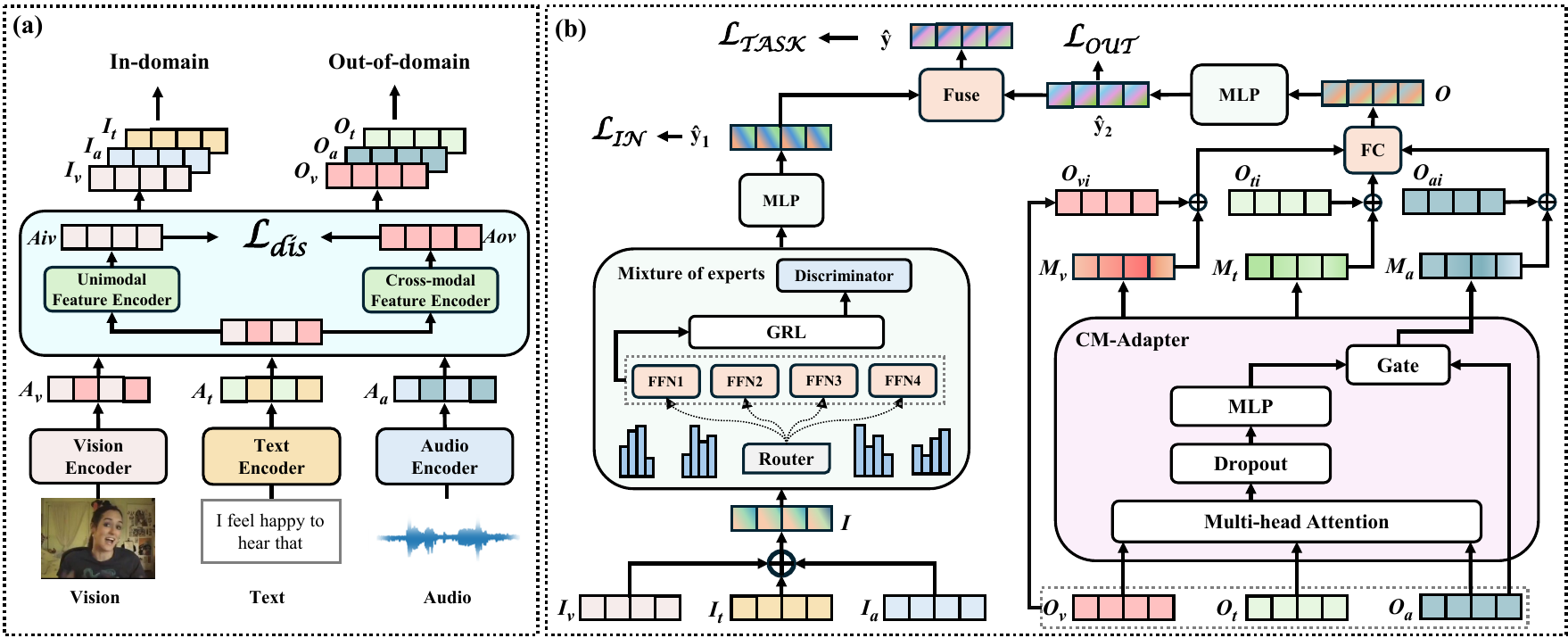}}
\caption{Framework of MIDG. \textbf{(a)}:Data preparation pipeline, using the entropy decoupling module. \textbf{(b)}: Pipeline of MSA task, divided into in-domain data flows and out-of-domain data flows. The final output of the model is obtained by performing a weighted sum of the results from the two processes.}
\label{figure}
\end{figure*}

\subsection{Domain Generalization}
\label{ssec:subhead}

The goal of domain generalization is to train models on a source domain so they can generalize to an unknown target domain without accessing to any target domain samples. Existing domain generalization methods\cite{oodsurvey} primarily include causal reasoning\cite{clue}, domain-invariant feature learning\cite{learning_in_order}, and model ensemble techniques. Among strategies for learning invariant features\cite{inv}, Domain Adversarial Training\cite{adv} is a common approach. It introduces a domain discriminator and employs adversarial learning to force the feature extractor to generate features that confuse the discriminator.

Recent research has begun exploring multimodal domain generalization, such as leveraging inter-modal complementarity to enhance generalization capabilities , or addressing cross-domain challenges by decoupling domain-invariant and modality-specific features .

\section{methodology}
\label{sec:pagestyle}
\subsection{Overall Architecture}
\label{ssec:subhead}

In this section, we will describe the overall architecture of the model. Figure\ref{figure} (a) illustrates our approach to preprocessing the data. First, we encode primary features from text, video and audio using encoders specific to each modality. Next, an information entropy decoupling module separates these features into two distinct distributions. As shown in Figure\ref{figure} (b), our model comprises two main components. One distribution is used as an in-domain training dataset and a multi-expert model is employed to enhance multimodal fusion for learning the MSA regression task. We treat the other distribution as out-of-domain data and use multimodal knowledge injection techniques to improve representation learning. The overall loss function comprises the losses from both pathways, thereby strengthening the model's ability to generalize during training.

\subsection{Data Preparation}
\label{ssec:subhead}


In MSA, the data type being processed is video, requiring steps such as feature extraction and encoding of information before inputting into the model. We define the encoded feature as $A_m$ ($m \in \{t, a, v\}$, $t$, $a$, and $v$ denotes text, audio, and vision respectively). To enhance the model's generalization capability and achieve efficient performance on out-of-domain data, we employ an information entropy decoupling module to explicitly decouple the unimodal representations. We first sequentially pass the unimodal features through two encoders, representing them as random variables $Aim$ and $Aom$. We define the mutual information $I(Aim, Aom)$ \cite{mutual}between these two random variables to simulate their correlation. To separate the distance of two representations, we formulate the problem as minimizing mutual information to achieve decoupling. Ultimately, we calculate the minimization result of the variational upper bound of mutual information as the optimization objective. The information entropy decoupling loss $\mathcal{L}_{DIS}$ can be expressed as follows:
\begin{align}
\mathcal{L}_{DIS} &=I\left(\boldsymbol{A}_{im}; \boldsymbol{A}_{om}\right) \\
I\left(\boldsymbol{A}_{im} ;\boldsymbol{A}_{om} \right)\leq  &D_{\mathrm{KL}}\left(q\left(\boldsymbol{A}_{im}\mid\boldsymbol{A}_{m}\right)\|p\left(\boldsymbol{A}_{om}\right)\right) \notag \\
&+D_{\mathrm{KL}}\left(q\left(\boldsymbol{A}_{om}\mid\boldsymbol{A}_{m}\right)\|p\left(\boldsymbol{A}_{im}\right)\right)  \\
&-\boldsymbol{E}  
\notag
\end{align}
Where $I(\cdot ; \cdot)$ means mutual information, $D_{\mathrm{KL}}$ means Kullback-Leibler divergence, $p$ means probability, $q$ means posterior distribution, $\boldsymbol{E}$ can be calculated as:
\begin{align}
\begin{aligned}
\boldsymbol{E}=\mathbb{E}_{q\left(\boldsymbol{A}_{im}|\boldsymbol{A}_{m}\right)q\left(\boldsymbol{A}_{om}|\boldsymbol{A}_{m}\right)} \left[\log p\left(\boldsymbol{A}_{m}\mid\boldsymbol{A}_{im},\boldsymbol{A}_{om}\right)\right]
\end{aligned}
\end{align}

After decoupling, features that conform to two distinct distributions are obtained: one corresponds to the source domain features, while the other is regarded as features from out-of-domain. The source-domain-specific feature representations, considered as data within the source domain, are directed to the in-domain processing pipeline, whereas the features following the other distribution, deemed as data from the target domain, are fed into the out-of-domain processing pipeline.

\subsection{Mixture of Invariant Experts (MoIE)}
\label{ssec:subhead}

For the in-domain data training pipeline, we first concatenate the representations from the three modalities to achieve a preliminary fusion. For the fused multimodal representation, unlike conventional approaches, we employ a Mixture of Invariant Experts (MoIE) architecture to replace the traditional fusion network. Its core objective is to extract highly discriminative and domain-agnostic features, that is, invariant features, from the in-domain data. The fused multimodal features are fed into this MoIE, and the output is a sentiment representation vector. This vector is then passed through a Multilayer Perceptron (MLP) to obtain the final sentiment prediction $\hat{y}_1$.

This Mixture of Invariant Experts consists of a router network, multiple experts, a Gradient Reversal Layer (GRL), and a discriminator. Specifically, we configure $K$ experts $\{E_1, E_2, \dots, E_K\}$. Each expert $E_k$ is a Feedforward Neural Network explicitly designed as a domain-invariant feature extractor. The outputs of all experts are aggregated via a weighted sum to produce the final output.

The router network $G$, made up of multiple linear layers and a Softmax scaling function, computes a weight distribution $\mathbf{g}(x) = [g_1(x), g_2(x), \dots, g_K(x)]$ conditioned on the semantic information of the fused multimodal vector input $\mathbf{I} = [I_t, I_a, I_v]$. Here, $g_k(x)$ denotes the probability that input $x$ is assigned to expert $k$. Based on this distribution, the router dynamically allocates tasks to different experts, enabling the model to adaptively select the most relevant expert(s) according to the input.

A Gradient Reversal Layer (GRL) is inserted between the experts and the discriminator within the MoIE. It receives the outputs from the experts and facilitates adversarial training between the experts and the discriminator. During backpropagation, the GRL reverses the gradients, thereby enabling adversarial loss optimization. The discriminator is a simple classifier implemented as a two-layer perceptron. To form an adversarial game with the experts, it compares the experts' output vectors with the true labels and is trained accordingly. Simultaneously, via the GRL, experts receive reversed gradients from the discriminator, aiming to maximize the loss function, which opposing the discriminator's goal of minimizing it. This adversarial process eventually forces the output vectors to converge to domain-invariant representations, achieving the overall goal of extracting invariant features. The loss function can be expressed as:
\begin{equation}
\begin{aligned}
\mathcal{L}_{in} = - \frac{1}{N} \sum_{i=1}^{N} \left[ d_i \cdot \log(D(E(\mathbf{I}^{(i)})))  \right.
\\ \phantom{=\;\;}
\left.+(1-d_i) \cdot \log(1 - D(E(\mathbf{I}^{(i)}))) \right]
\end{aligned}
\end{equation}
Where $d_i$ means the true domain label for the $i$-th sample, $D(\cdot)$ means the output of the Discriminator, $E$ denotes the expectation.
\subsection{Cross-Modal Adapter}
\label{ssec:subhead}

For the modal features of out-of-domain data, we directly feed the unimodal representations $O_m$ ($m \in \{t, a, v\}$) into the Cross-Modal Adapter. This module utilizes out-of-domain data to enhance modal representations through dynamic knowledge injection. It incorporates a multi-head attention mechanism. Specifically, for a target modality, we project it as the $\textbf{Q}$, while the remaining two modalities are projected as the $\textbf{K}$ and $\textbf{V}$, respectively. The formulation is as follows:
\begin{equation}
\text{Attention}(Q, K, V) = \text{Softmax}\left(\frac{QK^T}{\sqrt{d_k}}\right)V
\end{equation}

This computation achieves the knowledge injection into the target modality. The resulting attention output is then passed through a dropout layer and a MLP to obtain the initially enhanced features, which represent the knowledge we aim to inject. We introduce a learnable gated network, controlled by the target modality, to ensure that the injected features are complementary to the target modality. This gate dynamically balances the importance of knowledge originating from different source modalities.

For the final multimodal injection features $\{M_t, M_a, M_v\}$, we fuse them with the corresponding target modality to obtain the knowledge-injected feature-specific representations $\{O_{ti}, O_{ai}, O_{vi}\}$, which contain richer and more complete semantic information. These enhanced unimodal representations are then concatenated via a fully connected layer, producing the multimodal output $O$. Finally, $O$ is fed into an MLP to yield the out-of-domain sentiment prediction result $\hat{y}_2$.The loss function can be expressed as:
\begin{equation}
\mathcal{L}_{out} = \frac{1}{n} \| \mathbf{y} - \hat{\mathbf{y}_{2}} \|_2^2
\end{equation}

Where $n$ denotes the batch size, and $\mathbf{y}$ represents the true label value. The predictions from both the in-domain pipeline ($\hat{y}_1$) and the out-of-domain pipeline ($\hat{y}_2$) are combined via a weighted summation to produce the final prediction. This is supervised under the total task loss function $\mathcal{L}_{task}$, where $\alpha$ and $\beta$ are hyperparameters:
\begin{equation}
\mathcal{L}_{task} = \alpha \cdot \mathcal{L}_{in} + \beta \cdot \mathcal{L}_{out}
\end{equation}

\begin{table*}[htbp]
\centering
\normalsize
\setlength{\tabcolsep}{4pt} 
\renewcommand{\arraystretch}{1.1} 
\caption{\label{1}
    Domain generalization performance of model MIDG compared to SOTA approaches. }
\label{tab:model_results}
\begin{tabular}{l*{15}{c}}
\toprule
\multirow{2}{*}{Datasets} & \multicolumn{3}{c}{MOSI $\xrightarrow{}$ MOSEI} & \multicolumn{3}{c}{MOSI $\xrightarrow{}$ SIMS} & \multicolumn{3}{c}{MOESI $\xrightarrow{}$ SIMS} & \multicolumn{3}{c}{SIMS $\xrightarrow{}$ MOSI} & \multicolumn{3}{c}{SIMS $\xrightarrow{}$ MOSEI} \\
\cmidrule(lr){2-4} \cmidrule(lr){5-7} \cmidrule(lr){8-10} \cmidrule(lr){11-13} \cmidrule(lr){14-16}
 & ACC & F1 & MAE & ACC & F1 & MAE & ACC & F1 & MAE & ACC & F1 & MAE & ACC & F1 & MAE \\
\midrule
MuLT & 62.85 & 77.19 & 0.8891 & 47.21 & 42.35 & 0.6692 & 53.28 & 47.53 & 0.6642 & 54.57 & 43.49 & 0.8267 & 48.12 & 45.59 & 0.6598 \\
MISA & 66.47 & 76.52 & 0.7734 & 48.18 & 44.97 & 0.6484 & 55.19 & 48.75 & 0.6158 & 55.38 & 44.27 & 0.8236 & 49.45 & 47.37 & 0.6233 \\
Self-MM & 72.15 & 78.20 & 0.7453 & 48.65 & 43.29 & 0.6567 &  55.48& 49.96 & 0.6273 & 57.18 & 46.76 & 0.8242 & 51.59 & 47.62 & 0.6194 \\
ALMT & 80.19 & 80.09 & 0.6869 & 50.75 & 43.52 & 0.6491 & 56.67 & 49.78 & 0.5912 & 59.62 & 44.53 & 0.8071 & 51.01 & 44.74 & 0.6007 \\
CASP & 82.59 & 81.27 & 0.6987 & 50.89 & 51.04 & 0.6281 & 57.16 & 51.24 & 0.6107 & 58.14 & 50.94 & 0.8012 & 52.74 & 53.08 & 0.6084 \\
S\textsuperscript{2}LIF & 81.06 & 82.54 & 0.7015 & 51.46 & 50.19 & 0.6295 & 57.06 & \bf52.64 & 0.6128 & 58.46 & 51.27 & \bf0.7918 & 53.67 & 52.96 & 0.6102 \\
\midrule
\textbf{MIDG} & \bf83.97 & \bf83.95 & \bf0.6869 & \bf52.30 & \bf51.33 & \bf0.6089 & \bf58.64 & 51.58 & \bf0.5860 & \bf59.77 & \bf53.46 & 0.7975 & \bf55.66 & \bf55.01 & \bf0.5961 \\
\bottomrule
\end{tabular}
\end{table*}

\section{experiments}
\label{sec:typestyle}

\subsection{Datasets and Metrics}
\label{ssec:subhead}

We conducted experiments on three widely used multimodal datasets: CMU-MOSI\cite{zadeh2016mosi}, CMU-MOSEI\cite{bagher-zadeh-etal-2018-multimodal}, and CH-SIMS\cite{yu-etal-2020-ch}. The two English datasets, MOSI and MOSEI, comprise videos sourced from YouTube, with a duration ranging from 2 to 5 minutes. Videos are annotated with real number labels within the range [-3,3], where the sign of the value indicates sentimental polarity and the magnitude represents sentimental intensity. The Chinese dataset CH-SIMS comprises 2,281 video clips sourced from various platforms, with each sample manually annotated with an sentimental score ranging [-1,1]. In order to ensure fairness, metrics for both regression and classification tasks are employed: The mean absolute error (MAE), binary classification accuracy (ACC), and F1 score are the metrics of primary interest.

\begin{table}[htbp]
\centering
\setlength{\tabcolsep}{5pt} 
\renewcommand{\arraystretch}{1} 
\caption{Ablation results of MIDG}
\label{t2}
\begin{tabular}{cccccc}
\toprule
\multirow{2}{*}{MoIE} & \multirow{2}{*}{CM-Adapter} & \multicolumn{4}{c}{MOSI $\xrightarrow{}$ MOSEI} \\
\cmidrule(lr){3-6}
 &&ACC & Corr & F1 & MAE \\
\midrule
 \ding{56} & \ding{56} & 77.69 & 0.1596 & 0.7986 & 0.7257 \\
 \ding{52} & \ding{56} & 80.91 & 0.3684 & 0.7948 & 0.6941 \\
 \ding{56} & \ding{52} & 79.58 & 0.4379 & 0.8065 & 0.7142 \\
 \ding{52} & \ding{52} & \bf 83.97 & \bf0.5123 & \bf0.8395 & \bf0.6869 \\

\midrule
\multicolumn{2}{c}{\multirow{2}{*}{Modalities}} & \multirow{2}{*}{ACC} & \multirow{2}{*}{Corr} & \multirow{2}{*}{F1} & \multirow{2}{*}{MAE} \\ 
 \\
\midrule
\multicolumn{2}{c}{Text} & 81.48 & 0.768 & 82.69 & 0.8138 \\
\multicolumn{2}{c}{Audio} & 68.32 & 0.375 & 71.87 &  1.2538 \\
\multicolumn{2}{c}{Vision} & 70.61 & 0.461 & 72.24 & 0.9615 \\
\multicolumn{2}{c}{Text+Audio} & 84.96 & 0.784 & 84.35 & 0.7249 \\
\multicolumn{2}{c}{Text+Vision} & 83.27 & 0.779 & 84.82 & 0.7153 \\
\multicolumn{2}{c}{Text+Audio+Vision} & \bf87.94 & \bf0.812 & \bf86.16 & \bf0.6725 \\
\bottomrule
\end{tabular}
\end{table}
\subsection{Baselines}
\label{ssec:subhead}

Extensive domain generalization experiments were conducted on advanced baseline models for MSA, with the results being compared with those from model MIDG. For the purposes of this study, classic models were selected for MSA tasks: MuLT\cite{mult}, MISA\cite{hazarika2020misa}, Self-MM\cite{selfmm}, and ALMT\cite{almt}; moreover, we incorporated models that enhance domain generalization performance in MSA: CASP\cite{CASP} and S\textsuperscript{2}LIF\cite{learning_in_order}.

\subsection{Results}
\label{ssec:subhead}

During testing, we ensured that the data from the test set was processed only through the out-of-domain pipeline, which incorporating the CM-Adapter. Table \ref{tab:model_results} presents the domain generalization experiments of model MIDG on three datasets, along with performance comparisons with various methods. We reproduced these baselines under the same experimental conditions as ours, and the results demonstrate that our model outperforms these baselines. This indicates that training with simulated out-of-domain data in conjunction with in-domain data can lead to superior model performance. CASP employs contrastive adaptation and pseudo-label generation to accomplish domain adaptation, yet our model surpasses it by approximately 5.5\% in ACC on the SIMS→MOSEI generalization task. S\textsuperscript{2}LIF utilizes sequential learning strategy, our model achieves higher ACC across all five tasks. Furthermore, among the cross-dataset domain generalization experiments, the lowest performance across metrics was observed in MOSI→SIMS, suggesting that cross-lingual domain generalization remains a challenging area requiring further improvement.

In addition to domain generalization experiments, we also evaluated the model on conventional Multimodal Sentiment Analysis (MSA) tasks using the MOSI and MOSEI datasets. We compared the experimental results of our model with four established baselines: MuLT, MISA, Self-MM, and ALMT. And we adopted four metrics for comprehensive performance comparison: binary accuracy (ACC), correlation coefficient (Corr), F1-score, and mean absolute error (MAE). As shown in Table \ref{t3}, on the MOSEI dataset, our model achieved the best results across all four metrics. On the MOSI dataset, it outperformed the baselines in ACC, Corr, and MAE.

\begin{table}[htbp]
\centering
\setlength{\tabcolsep}{2pt} 
\renewcommand{\arraystretch}{1.1} 
\caption{Experimental results on the CMU-MOSI and CMU-MOSEI datasets.}
\label{t3}
\begin{tabular}{ccccccccc}
\toprule
\multirow{2}{*}{Datasets}  & \multicolumn{4}{c}{MOSI} & \multicolumn{4}{c}{MOSEI} \\
\cmidrule(lr){2-5} \cmidrule(lr){6-9}
 &ACC & Corr & F1 & MAE &ACC & Corr & F1 & MAE \\
\midrule
 MuLT & 81.16 & 0.711 & 81.04 & 0.8891 & 81.64 & 0.703 & 81.56 & 0.591 \\
 MISA & 82.57 & 0.761 & 82.54 & 0.7830 & 85.52 & 0.765 & 85.37 & 0.555 \\
 Self-MM & 85.98 & 0.798 & 85.95 & 0.713 & 85.17 & 0.765 & 85.30 & 0.530\\
 ALMT & 86.43 & 0.805 & \bf86.47 & 0.6830 & 86.79 & 0.779 & 85.15 & 0.526 \\
 \midrule
 MIDG & \bf87.94 & \bf0.812 & 86.16 & \bf0.6725 & \bf87.96 & \bf0.781 & \bf85.43 & \bf0.509 \\
\bottomrule
\end{tabular}
\end{table}
\subsection{Ablation Study}
\label{ssec:subhead}

We conducted ablation studies by sequentially removing key modules from the framework, as summarized in Table \ref{t2}. When both the Mixture of Invariant Experts (MoIE) module and the Cross-Modal Adapter were removed, all metrics exhibited the most significant decline, with the most pronounced drop observed in ACC (a decrease of 7.48\%). This reduction far exceeded those observed when removing either MoIE or CM-Adapter alone, indicating that these modules synergistically enhance the model’s ability to predict sentiment polarity.

Furthermore, we conducted modal ablation experiments on the MOSI dataset. The results demonstrated that the model performed significantly worse with unimodal data compared to multimodal data, with the textual modality achieving the best performance among the three unimodal scenarios. This confirms that the textual modality carries the most abundant semantic features, and that multimodal data can convey richer semantic information compared to unimodal data, thereby facilitating sentiment analysis.

\section{conclusion}
\label{sec:majhead}

We propose a novel framework MIDG for domain generalization in Multimodal Sentiment Analysis. It incorporates a Mixture of Invariant Experts and a Cross-Modal Adapter. The former is designed to extract domain-invariant features, while the latter enhances multimodal representations by knowledge injection. These modules are systematically trained using both in-domain and out-of-domain data, which are simulated via an Information Entropy Disentanglement module to follow two distinct distributions. Extensive experiments conducted on three datasets demonstrate the superior generalization capability of MIDG in sentiment prediction.


\bibliographystyle{IEEEbib}
\bibliography{refs}

@article{mutual,
  title={Multivariate information transmission},
  author={McGill, William},
  journal={Transactions of the IRE Professional Group on Information Theory},
  volume={4},
  number={4},
  pages={93--111},
  year={1954},
  publisher={IEEE}
}

@ARTICLE{inv,
       author = {{Arjovsky}, Martin and {Bottou}, L{\'e}on and {Gulrajani}, Ishaan and {Lopez-Paz}, David},
        title = "{Invariant Risk Minimization}",
      journal = {arXiv e-prints},
         year = 2019,
        month = jul,
          eid = {arXiv:1907.02893},
        pages = {arXiv:1907.02893},
          doi = {10.48550/arXiv.1907.02893},
archivePrefix = {arXiv},
       eprint = {1907.02893},
 primaryClass = {stat.ML},
       adsurl = {https://ui.adsabs.harvard.edu/abs/2019arXiv190702893A}
}

@ARTICLE{survey2,
       author = {{Wei}, Xiaokai and {Wang}, Shen and {Zhang}, Dejiao and {Bhatia}, Parminder and {Arnold}, Andrew},
        title = "{Knowledge Enhanced Pretrained Language Models: A Compreshensive Survey}",
      journal = {arXiv e-prints},
         year = 2021,
        month = oct,
          eid = {arXiv:2110.08455},
        pages = {arXiv:2110.08455},
          doi = {10.48550/arXiv.2110.08455},
archivePrefix = {arXiv},
       eprint = {2110.08455},
 primaryClass = {cs.CL},
       adsurl = {https://ui.adsabs.harvard.edu/abs/2021arXiv211008455W}
}

@inproceedings{fuse,
    title = "Efficient Low-rank Multimodal Fusion With Modality-Specific Factors",
    author = "Liu, Zhun  and
      Shen, Ying  and
      Lakshminarasimhan, Varun Bharadhwaj  and
      Liang, Paul Pu  and
      Bagher Zadeh, AmirAli  and
      Morency, Louis-Philippe",
    editor = "Gurevych, Iryna  and
      Miyao, Yusuke",
    booktitle = "Proceedings of the 56th Annual Meeting of the Association for Computational Linguistics (Volume 1: Long Papers)",
    month = jul,
    year = "2018",
    address = "Melbourne, Australia",
    publisher = "Association for Computational Linguistics",
    url = "https://aclanthology.org/P18-1209/",
    doi = "10.18653/v1/P18-1209",
    pages = "2247--2256"
}

@inproceedings{confede,
    title = "{C}on{FEDE}: Contrastive Feature Decomposition for Multimodal Sentiment Analysis",
    author = "Yang, Jiuding  and
      Yu, Yakun  and
      Niu, Di  and
      Guo, Weidong  and
      Xu, Yu",
    editor = "Rogers, Anna  and
      Boyd-Graber, Jordan  and
      Okazaki, Naoaki",
    booktitle = "Proceedings of the 61st Annual Meeting of the Association for Computational Linguistics (Volume 1: Long Papers)",
    month = jul,
    year = "2023",
    address = "Toronto, Canada",
    publisher = "Association for Computational Linguistics",
    url = "https://aclanthology.org/2023.acl-long.421/",
    doi = "10.18653/v1/2023.acl-long.421",
    pages = "7617--7630"
}

@ARTICLE{adv,
       author = {{Albuquerque}, Isabela and {Monteiro}, Jo{\~a}o and {Darvishi}, Mohammad and {Falk}, Tiago H. and {Mitliagkas}, Ioannis},
        title = "{Generalizing to unseen domains via distribution matching}",
      journal = {arXiv e-prints},
         year = 2019,
        month = nov,
          eid = {arXiv:1911.00804},
        pages = {arXiv:1911.00804},
          doi = {10.48550/arXiv.1911.00804},
archivePrefix = {arXiv},
       eprint = {1911.00804},
 primaryClass = {cs.LG},
       adsurl = {https://ui.adsabs.harvard.edu/abs/2019arXiv191100804A}
}

@ARTICLE{oodsurvey,
       author = {{Liu}, Jiashuo and {Shen}, Zheyan and {He}, Yue and {Zhang}, Xingxuan and {Xu}, Renzhe and {Yu}, Han and {Cui}, Peng},
        title = "{Towards Out-Of-Distribution Generalization: A Survey}",
      journal = {arXiv e-prints},
         year = 2021,
        month = aug,
          eid = {arXiv:2108.13624},
        pages = {arXiv:2108.13624},
          doi = {10.48550/arXiv.2108.13624},
archivePrefix = {arXiv},
       eprint = {2108.13624},
 primaryClass = {cs.LG},
       adsurl = {https://ui.adsabs.harvard.edu/abs/2021arXiv210813624L}
}

@ARTICLE{clue,
       author = {{Sun}, Teng and {Wang}, Wenjie and {Jing}, Liqiang and {Cui}, Yiran and {Song}, Xuemeng and {Nie}, Liqiang},
        title = "{Counterfactual Reasoning for Out-of-distribution Multimodal Sentiment Analysis}",
      journal = {arXiv e-prints},
         year = 2022,
        month = jul,
          eid = {arXiv:2207.11652},
        pages = {arXiv:2207.11652},
          doi = {10.48550/arXiv.2207.11652},
archivePrefix = {arXiv},
       eprint = {2207.11652},
 primaryClass = {cs.CL},
       adsurl = {https://ui.adsabs.harvard.edu/abs/2022arXiv220711652S}
}

@ARTICLE{selfmm,
       author = {{Yu}, Wenmeng and {Xu}, Hua and {Yuan}, Ziqi and {Wu}, Jiele},
        title = "{Learning Modality-Specific Representations with Self-Supervised Multi-Task Learning for Multimodal Sentiment Analysis}",
      journal = {arXiv e-prints},
         year = 2021,
        month = feb,
          eid = {arXiv:2102.04830},
        pages = {arXiv:2102.04830},
          doi = {10.48550/arXiv.2102.04830},
archivePrefix = {arXiv},
       eprint = {2102.04830},
 primaryClass = {cs.CL},
       adsurl = {https://ui.adsabs.harvard.edu/abs/2021arXiv210204830Y}
}

@ARTICLE{mult,
       author = {{Tsai}, Yao-Hung Hubert and {Bai}, Shaojie and {Liang}, Paul Pu and {Zico Kolter}, J. and {Morency}, Louis-Philippe and {Salakhutdinov}, Ruslan},
        title = "{Multimodal Transformer for Unaligned Multimodal Language Sequences}",
      journal = {arXiv e-prints},
         year = 2019,
        month = jun,
          eid = {arXiv:1906.00295},
        pages = {arXiv:1906.00295},
          doi = {10.48550/arXiv.1906.00295},
archivePrefix = {arXiv},
       eprint = {1906.00295},
 primaryClass = {cs.CL},
       adsurl = {https://ui.adsabs.harvard.edu/abs/2019arXiv190600295T}
}

@ARTICLE{CASP,
       author = {{Guo}, Zirun and {Jin}, Tao and {Xu}, Wenlong and {Lin}, Wang and {Wu}, Yangyang},
        title = "{Bridging the Gap for Test-Time Multimodal Sentiment Analysis}",
      journal = {arXiv e-prints},
         year = 2024,
        month = dec,
          eid = {arXiv:2412.07121},
        pages = {arXiv:2412.07121},
          doi = {10.48550/arXiv.2412.07121},
archivePrefix = {arXiv},
       eprint = {2412.07121},
 primaryClass = {cs.LG},
       adsurl = {https://ui.adsabs.harvard.edu/abs/2024arXiv241207121G}
}

@ARTICLE{learning_in_order,
       author = {{Zhao}, Xianbing and {Qu}, Lizhen and {Feng}, Tao and {Cai}, Jianfei and {Tang}, Buzhou},
        title = "{Learning in Order! A Sequential Strategy to Learn Invariant Features for Multimodal Sentiment Analysis}",
      journal = {arXiv e-prints},
         year = 2024,
        month = sep,
          eid = {arXiv:2409.04473},
        pages = {arXiv:2409.04473},
          doi = {10.48550/arXiv.2409.04473},
archivePrefix = {arXiv},
       eprint = {2409.04473},
 primaryClass = {cs.LG},
       adsurl = {https://ui.adsabs.harvard.edu/abs/2024arXiv240904473Z}
}

@inproceedings{almt,
    title = "Learning Language-guided Adaptive Hyper-modality Representation for Multimodal Sentiment Analysis",
    author = "Zhang, Haoyu  and
              Wang, Yu  and
              Yin, Guanghao  and
              Liu, Kejun  and
              Liu, Yuanyuan  and
              Yu, Tianshu",
    booktitle = "Proceedings of the 2023 Conference on Empirical Methods in Natural Language Processing",
    year = "2023",
    publisher = "Association for Computational Linguistics",
    pages = "756--767"
}

@article{yu2023conki,
  title={ConKI: Contrastive knowledge injection for multimodal sentiment analysis},
  author={Yu, Yakun and Zhao, Mingjun and Qi, Shi-ang and Sun, Feiran and Wang, Baoxun and Guo, Weidong and Wang, Xiaoli and Yang, Lei and Niu, Di},
  journal={arXiv preprint arXiv:2306.15796},
  year={2023}
}

@article{hazarika2020misa,
  title={MISA: Modality-Invariant and-Specific Representations for Multimodal Sentiment Analysis},
  author={Hazarika, Devamanyu and Zimmermann, Roger and Poria, Soujanya},
  journal={arXiv preprint arXiv:2005.03545},
  year={2020}
}

@article{zadeh2016mosi,
  title={Mosi: multimodal corpus of sentiment intensity and subjectivity analysis in online opinion videos},
  author={Zadeh, Amir and Zellers, Rowan and Pincus, Eli and Morency, Louis-Philippe},
  journal={arXiv preprint arXiv:1606.06259},
  year={2016}
}

@inproceedings{bagher-zadeh-etal-2018-multimodal,
    title = "Multimodal Language Analysis in the Wild: {CMU}-{MOSEI} Dataset and Interpretable Dynamic Fusion Graph",
    author = "Bagher Zadeh, AmirAli  and
      Liang, Paul Pu  and
      Poria, Soujanya  and
      Cambria, Erik  and
      Morency, Louis-Philippe",
    editor = "Gurevych, Iryna  and
      Miyao, Yusuke",
    booktitle = "Proceedings of the 56th Annual Meeting of the Association for Computational Linguistics (Volume 1: Long Papers)",
    month = jul,
    year = "2018",
    address = "Melbourne, Australia",
    publisher = "Association for Computational Linguistics",
    url = "https://aclanthology.org/P18-1208",
    doi = "10.18653/v1/P18-1208",
    pages = "2236--2246",
}

@inproceedings{yu-etal-2020-ch,
    title = "{CH}-{SIMS}: A {C}hinese Multimodal Sentiment Analysis Dataset with Fine-grained Annotation of Modality",
    author = "Yu, Wenmeng  and
      Xu, Hua  and
      Meng, Fanyang  and
      Zhu, Yilin  and
      Ma, Yixiao  and
      Wu, Jiele  and
      Zou, Jiyun  and
      Yang, Kaicheng",
    editor = "Jurafsky, Dan  and
      Chai, Joyce  and
      Schluter, Natalie  and
      Tetreault, Joel",
    booktitle = "Proceedings of the 58th Annual Meeting of the Association for Computational Linguistics",
    month = jul,
    year = "2020",
    address = "Online",
    publisher = "Association for Computational Linguistics",
    url = "https://aclanthology.org/2020.acl-main.343",
    doi = "10.18653/v1/2020.acl-main.343",
    pages = "3718--3727",
}

@ARTICLE{bert,
       author = {{Devlin}, Jacob and {Chang}, Ming-Wei and {Lee}, Kenton and {Toutanova}, Kristina},
        title = "{BERT: Pre-training of Deep Bidirectional Transformers for Language Understanding}",
      journal = {arXiv e-prints},
         year = 2018,
        month = oct,
          eid = {arXiv:1810.04805},
        pages = {arXiv:1810.04805},
          doi = {10.48550/arXiv.1810.04805},
archivePrefix = {arXiv},
       eprint = {1810.04805},
 primaryClass = {cs.CL},
       adsurl = {https://ui.adsabs.harvard.edu/abs/2018arXiv181004805D}
}

\end{document}